\documentclass{ecai}
\usepackage{times}
\usepackage{graphicx}
\usepackage{latexsym}

\usepackage{color,xcolor}
\usepackage{amsmath}
\usepackage{booktabs}
\usepackage{amssymb}
\usepackage{amsfonts}
\usepackage{mathrsfs}
\usepackage{subfigure}
\usepackage{enumitem}
\usepackage[misc]{ifsym}

\begin{document}

\title{Assessing the Memory Ability of Recurrent Neural Networks}

\author{Cheng Zhang\institute{Tianjin Key Laboratory of Cognitive Computing and Application, Tianjin University, Tianjin, China, email: zccode@gmail.com} \and Qiuchi Li\institute{Department of Information Engineering, University of Padua, Padua, Italy, email: qiuchili@dei.unipd.it} \and Lingyu Hua\institute{School of Computer Science, Beijing Institute of Technology, Beijing, China, email: hualingyu@hotmail.com} \and Dawei Song\institute{School of Computer Science, Beijing Institute of Technology, Beijing, China, email: dwsong@bit.edu.cn (Corresponding Author)}}



\maketitle
\bibliographystyle{ecai}

\begin{abstract}
It is known that Recurrent Neural Networks (RNNs) can remember, in their hidden layers, part of the semantic information expressed by a sequence (e.g., a sentence) that is being processed. Different types of recurrent units have been designed to enable RNNs to remember information over longer time spans. However, the memory abilities of different recurrent units are still theoretically and empirically unclear, thus limiting the development of more effective and explainable RNNs. To tackle the problem, in this paper, we identify and analyze the internal and external factors that affect the memory ability of RNNs, and propose a Semantic Euclidean Space to represent the semantics expressed by a sequence. Based on the Semantic Euclidean Space, a series of evaluation indicators are defined to measure the memory abilities of different recurrent units and analyze their limitations. These evaluation indicators also provide a useful guidance to select suitable sequence lengths for different RNNs during training.
\end{abstract}

\section{Introduction}

Recurrent Neural Networks (RNNs) have been widely applied in natural language processing tasks and have shown a proven ability to process text sequences such as sentences. The effectiveness of RNNs is attributed to their exquisite network structure and the reuse of hidden layers, which enable RNNs to process text sequences of arbitrary length in theory, and memorize the semantics of text sequences encoded in the hidden layers. However, in practice the memory ability of an RNN is limited by factors that are external or internal to the network structure.

\begin{itemize}[leftmargin=*]

\item \textbf{External Factor:} The length of input sequences is the external factor that affects the memory ability of an RNN. It is not related to the RNN's particular setting, but relies solely on the input data. For example, in order to solve the gradient vanishing and exploding problem \cite{vanishing-rnn} that occurs on excessively long sequences, we can manually shorten the length of input sequences to $T$ by various means, such as removing stop words, or using a Backpropagation Through Time (BPTT) strategy \cite{bptt} to limit the distance of the gradient back propagation to $T$. Consequently, the length of sequences that can be memorized by an RNN would be no greater than $T$.

\item \textbf{Internal Factor:} The particular type of recurrent unit employed by an RNN is the internal factor that affects the RNN's memory ability. For instance, it was difficult to capture long-term dependencies in a sequence with the early Vanilla Recurrent Neural Network (VRNN) \cite{rnn-language-model}, whereas the Long Short-Term Memory (LSTM) approach \cite{lstm-3gate} and Gated Recurrent Unit (GRU) \cite{gru} have improved the memory ability through well-designed gate structures.

\end{itemize}

Therefore, the external and internal factors determine the upper and lower bounds of an RNN's memory ability respectively. If a recurrent unit with limited memory ability (internal factor) is used to process a long sequence, the RNN's hidden layer will be internally constrained from effectively representing the semantics of the original sequence. It may also result in a reduced training speed. Likewise, a recurrent unit with a powerful memory ability will not be able to take its full advantage when it is used to process a short sequence (external factor).

Currently, both factors are empirically adjusted as hyper-parameters with respect to a specific task, but they may not always reflect the actual memory ability of the chosen recurrent unit. The lack of theoretical insights and effective ways to quantify the difference in memory ability between different types of recurrent units largely limits the rational use of them and the development of new recurrent units with a stronger memory ability. This paper aims to fill the gap by developing a principled approach to measuring and comparing the memory abilities of different recurrent units. 

Specifically, an RNN continually remembers the semantics of an input sequence in the hidden layer through a recurrent unit, and then performs a particular task based on the hidden layer. The hidden layer is vital, as it not only memorizes the semantics of the original sequence, but also further adjusts the semantics to suit the target task. Therefore, the internal memory ability of an RNN is reflected by the semantic difference between the original input sequence and the hidden layer sequence at the processing time. In this paper (Section \ref{sec:rnn}), we define a range of evaluation indicators of memory ability to quantitatively measure such semantic difference. Further, we analyze the relationship between the internal and external factors that affect the memory ability, by observing how the values of these indicators change with respect to varying input sequence length (as external indicator).

The theoretical foundation of the above analysis is how to represent the semantics of a text sequence. Semantics is concerned with the relationship between signifiers such as words, signs and symbols, and what they stand for in reality \cite{semantics_book}. Words are semantic units that convey meanings \cite{meaning}, and typically an $n$-dimensional word vector can be used to represent the semantics of a word. A sentence consists of a sequence of words. The semantics of a word sequence should contain not only the meanings of the individual words, but also the meanings emerging from different sub-sequences (word combinations) within the sequence. To this end, we propose, in Section \ref{sec:ses}, an $n$-dimensional Semantic Euclidean space (denoted as $\mathbb{SR}^n$), which is extended from an $n$-dimensional Euclidean space of word embeddings (denoted as $\mathbb{R}^n$), to clarify the mapping relationship between sequence semantics in $\mathbb{SR}^n$ and each point (word vector) in $\mathbb{R}^n$. Furthermore, we propose a novel representation of sequence semantics with a Convex Hull in $\mathbb{SR}^n$, which provides a powerful mathematical formulation and solid theoretical basis to assess the memory ability of RNNs. Finally, extensive experiments are carried out, for an in-depth theoretical and empirical analysis on the memory abilities of different types of recurrent units. The results would provide a practical guidance for setting the most appropriate length of input sequence that matches the memory ability of the recurrent unit used in an RNN.  

\section{Related Work}

The research on the internal mechanism of RNNs and comparison between different types of recurrent units has drawn wide attention. Earlier work analyzed various abilities of RNNs from a theoretical perspective, such as the universal approximation ability \cite{ua_rnn} and the generalization ability \cite{ga_rnn}, but neglected analyzing the memory ability of RNNs in natural language processing tasks. An empirical comparison of different types of recurrent units revealed that LSTM and GRU is more effective than VRNN on specific tasks~\cite{compare-rnns} , but the study failed to present the approximate sequence length that different types of recurrent units can memorize. Karpathy et al. (2015) explored the internal mechanism of LSTM on a microscopic level, and established a mapping between the neurons of hidden layers and the content represented \cite{lifeifei-lstm}. For example, one neuron can represent the beginning and end of a sentence, while another may indicate a punctuation mark. On a higher level, the internal mechanism of RNNs was analyzed by Hou and Zhou (2018) \cite{fsa} by modeling an RNN as a Finite State Automaton (FSA). Nonetheless, the problem of how much information an RNN can memorize and how this could be used to guide the setting of input sequence length, is yet to be solved. There have  also been works on evaluating the different abilities of LSTM, such as the ability of LSTM to learn syntax-sensitive dependencies \cite{tacl_assessing,enguehard2017exploring} and context-free grammars~\cite{sennhauser2018evaluating}. However, there is still a lack of horizontal comparison among different recurrent units in their memory ability.

\section{Measuring Sequence Semantics}
\label{sec:ses}

The widely used word embeddings are commonly assumed to be an $n$-dimensional Euclidean space ($\mathbb{R}^n$), in which the meaning of each word in a pre-defined vocabulary is represented by an $n$-dimensional vector. However, there is a difference between ``semantic meaningfulness'' and ``correspondence to an actual word''. For example, implicit or new meaning may emerge from a sequence of words. A vector representing such meaning may not necessarily associated to an exact word. Therefore, word embeddings only construct a small part of the relationship between word vectors and semantics, limited to a vocabulary of words. As a result, with $\mathbb{R}^n$ it is difficult to capture the implicit semantics that are not corresponding to any existing words in the vocabulary. For example, when dealing with the analogy problem $a:b \quad c:d$, for which  $x_d=x_b-x_a+x_c$ needs to be computed under a certain distance metric~\cite{mikolov2013linguistic}, there may not exist an appropriate word $d$ that corresponds to the derived vector representation $x_d$. Similarly, the vectors in the hidden layer of RNNs that are used to memorize the original sequence may hardly find their exact correspondence words. Thus using the word embedding space only would seem not enough to measure the hidden semantics of words and sequences. 

Therefore, we propose an $n$-dimensional Semantic Euclidean Space (denoted as $\mathbb{SR}^n$), which contains the semantic counterparts of all points in $\mathbb{R}^n$ and also allows the implicit semantics to be represented. A vector in the Semantic Euclidean Space is essentially a basic unit to express a latent semantics, and is therefore assigned with a semantic meaning. It may or may not be associated to the meaning of an actual word. Moreover, we posit that the semantics of a sequence can be represented as a convex hull in $\mathbb{SR}^n$, and design convex hull-based metrics as evaluation indicators for the memory ability of RNNs.

\begin{figure}[!tbp]
	\centering 
	\includegraphics[width=0.7\linewidth]{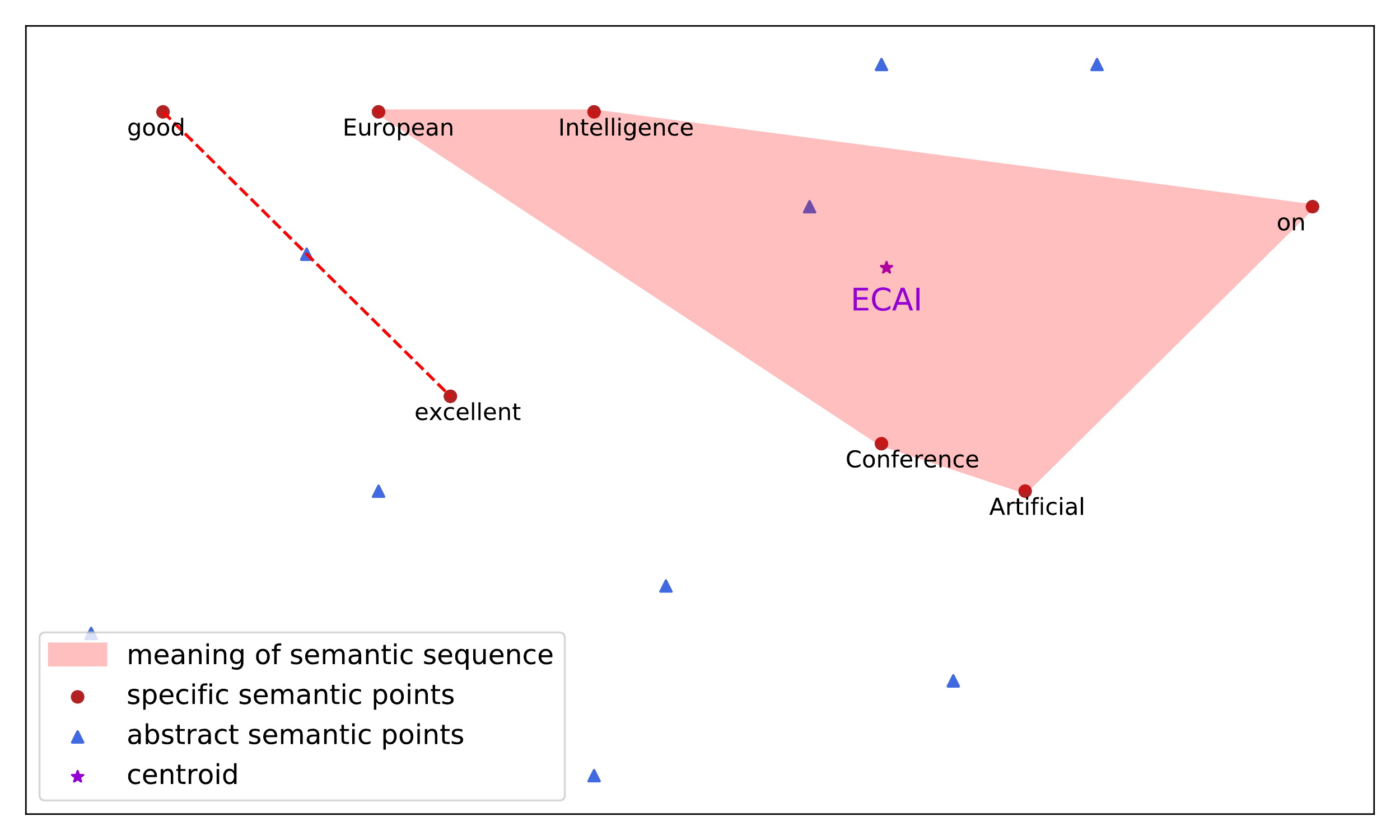}
	\caption{The sequence ``European Conference on Artificial Intelligence'' in Semantic Euclidean Space.}
	\label{fig:srn}
\end{figure}

\subsection{Definition of Semantic Euclidean Space}

The proposed $n$-dimensional Semantic Euclidean Space ($\mathbb{SR}^n$) is extended from the $n$-dimensional Euclidean space ($\mathbb{R}^n$). $\mathbb{R}^n$ contains a set of totally ordered arrays composed of $n$ real numbers, $n$ is a positive integer \cite{n-dimensional}:
	\begin{align}
		\mathbb{R}^n = \{x=(x_1, x_2, ..., x_n)|x_i\in\mathbb{R}, i=1,...,n\}
	\end{align}
In the $n$-dimensional Semantic Euclidean Space, each $n$-dimensional vector in $\mathbb{R}^n$ adopts a semantic correspondence:
\begin{align}
		\mathbb{SR}^n = \{&\forall x=(x_1, x_2, ..., x_n) \in\mathbb{R}^n | x\rightarrow \text{semantics}\}
\end{align}

Each point in $\mathbb{SR}^n$ is an $n$-dimensional vector and has a semantic meaning, hence called a \textit{semantic point}. Assuming that the size of a vocabulary is $V$, each word in the vocabulary is represented by an $n$-dimensional word vector in $\mathbb{R}^n$. If a semantic point explicitly corresponds with a word in the vocabulary, it is called a \textit{specific semantic point}. The difference set $\mathbb{SR}^n \setminus V^n$ denotes the set of semantic points that cannot be described by words in current vocabulary, called \textit{abstract semantic points}. Thus a set of pre-trained $n$-dimensional word vectors is actually a subset of $\mathbb{SR}^n$ containing only specific semantic points.

As is shown in Figure \ref{fig:srn}, the meanings represented by the specific semantic points can be described by words such as \emph{good} and \emph{excellent}, but we may not find any words in the vocabulary that correspond to the semantics carried by abstract semantic points. For example, the vector (abstract semantic point) on the dotted line connecting the vectors of \emph{good} and \emph{excellent} means \emph{good but not good enough to be excellent}, it is difficult to find the right word in the vocabulary to express this semantics. To represent the semantics behind an abstract semantic point, we need to use its nearest specific semantic point or invent a new word.

\subsection{Representing Sequence Semantics as Convex Hull}

A sentence consists of sequence of words. The semantics of a sentence, in addition to the semantics expressed by the individual words in the sentence, should also include the implicit semantics produced by possible combinations of words. This is consistent with the concept of convex hull in $\mathbb{R}^n$.  Therefore, the semantics of a sequence $\mathcal{X}=\{x_1, x_2, ..., x_n | x_i \in \mathbb{SR}^n\}$ can be expressed by the convex hull of all semantic points in $\mathcal{X}$.
\begin{align}
		\text{Conv}(\mathcal{X})= \left\{ \sum_{i=1}^{\left|\mathcal{X}\right|}\alpha_{i}x_{i} \bigg| 
		 \alpha_{i} \geq 0 \wedge
		\sum_{i=1}^{\left|\mathcal{X}\right|}\alpha_{i} = 1 \right\}
\end{align}

The convex hull of a finite point set $\mathcal{X}$ is the set of all convex combinations of its points \cite{convex_hull}. In a convex combination, each point $x_{i}$ in $\mathcal{X}$ is assigned with a weight or coefficient $\alpha_i$ in such a way that the coefficients are all non-negative and summed up to one. These weights are used in a weighted average of the points. For each choice of coefficients, the resulting convex combination is a unique point within the convex hull, and the whole convex hull can be formed by all possible choices of coefficients. Thus we can use convex hull to express the semantics of a sequence, which includes not only the meanings of all specific semantic points in the sequence, but also the implicit meanings (abstract semantic points) emerging from possible combinations of the specific semantic points.

It is interesting to show that the attention mechanism~\cite{attention} is equivalent to utilizing partial semantics of a sequence, which is a subset of the convex hull for a sequence. The core of the attention mechanism is to calculate a context vector $c$ from a set of sequences $h_1, h_2, ..., h_n$.
\begin{align}
c=\sum_{i=1}^{n}\alpha_{i}h_{i} \bigg| \sum_{i=1}^{n}\alpha_{i} = 1, \alpha_{i} \geq 0
\end{align}


Based on the discussions above, the context vector $c$ can be seen as a point in the convex hull of the sequence. A particular attention mechanism is potentially able to produce a subset of possible choices of $\alpha_i$ and hence give rise to a subset of convex hull.  In a target task, the optimal coefficients $\alpha_i$s are learned, and the most characteristic semantic point $c$ within the subset of convex hull is therefore identified. Different attention mechanisms or different weighting schemes may give rise to different convex combination weights and different points in the convex hull. In this way, the proposed convex hull delineates the set of sequence representations that could possibly be produced by an attention mechanism based on the word representations. Therefore, the convex hull is a reasonable choice for evaluating the representation capacity of the word representations in a sequence.

The semantic representation of a sequence corresponds to the meaning of the sequence, denoted as \textbf{ME}. For any sequence of semantic points, such as $K_m^n=\{k_m, k_{m+1}, ..., k_n | k_i \in \mathbb{SR}^n\}$, the meaning of $K_m^n$ can be modeled as the following formula:
\begin{align}
		\text{ME}(K_m^n) = &\text{Conv}(K_m^n)
\end{align}
For a visual illustration, the red shaded area in Figure \ref{fig:srn} is the convex hull that represents the meaning of a phrase \emph{"European Conference on Artificial Intelligence"}.

\subsection{Central Idea of a Sequence}

The meaning of a sequence should be centered around a central idea. For example, the central idea of \emph{``European Conference on Artificial Intelligence''} is \emph{``ECAI''}. As introduced above, ME scopes the meaning of a sequence as an area in $\mathbb{SR}^n$. Intuitively, the central idea of the sequence should be at the center of the ME area. Thus it is natural to use the centroid of convex hull to reflect the central idea of the sequence.

\begin{figure}[!tbp]
\centering
\includegraphics[width=0.8\linewidth]{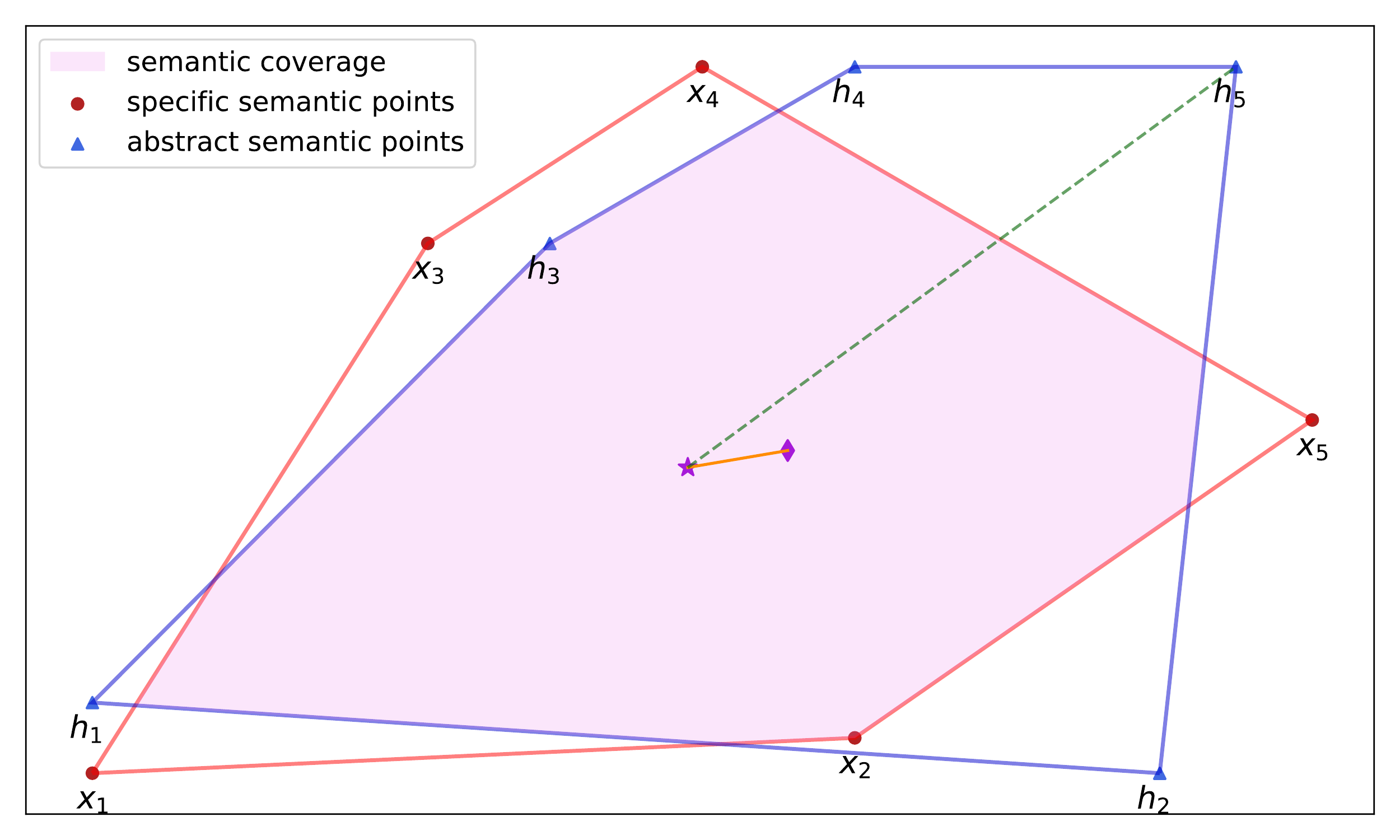}
\caption{The semantic difference between an origin sequence $x_1, x_2, x_3, x_4, x_5$ and the hidden layer sequence $h_1, h_2, h_3, h_4, h_5$ learned by RNNs. The star represents the central idea of the origin sequence, and the central idea of hidden layer sequence is represented by diamond.}
\label{fig:duo_convex}
\end{figure}

In $\mathbb{R}^n$, a centroid is the mean position of all the points in all of the coordinate directions. The centroid of a subset $\mathcal{X}$ of $\mathbb{R}^n$ is computed as follows:
\begin{align}
	\text{Centroid}(\mathcal{X}) = \frac{\int{xg(x)dx}}{\int{g(x)dx}}
\end{align}

where the integrals are taken over the whole space $\mathbb{R}^n$, and $g$ is the characteristic function of the subset, which is 1 inside $\mathcal{X}$ and 0 outside it \cite{centroid}.

Similarly, for a sequence $\mathcal{X} \subset \mathbb{SR}^n$, $\text{Centroid}(\mathcal{X})$ refers to a specific or abstract semantic point in $\mathbb{SR}^n$ that expresses the central idea of the sequence. For a specific sequence $K_m^n=\{k_m, k_{m+1}, ..., k_n | k_i \in  \mathbb{SR}^n\}$, the meaning of $K_m^n$ can be calculated by $\text{ME}(K_m^n) = \text{Conv}(K_m^n)$, so the central idea (denoted as \textbf{CI}) of $K_m^n$ is formulated as below:
\begin{align}
		\text{CI}(K_m^n)=&\text{Centroid}(\text{Conv}(K_m^n))
\end{align}

Note that the central idea of a sequence needs to be calculated as $\text{Centroid}(\text{Conv}(K_m^n))$, instead of $\text{Centroid}(K_m^n)$ directly. This definition guarantees that the central idea of a sequence lies within the convex hull (meaning) of the sequence. In contrast, even though the geometric centroid of a convex object always lies within the area representing its meaning, a non-convex object might have a centroid that is outside the area, which is undesirable. 

As shown in Figure \ref{fig:srn}, the central idea of \emph{``European Conference on Artificial Intelligence''} is represented by the purple star standing for \emph{``ECAI''}. When the vocabulary does not contain \emph{``ECAI''}, it is an abstract semantic point and we can express it with its nearest specific semantic point. Alternatively, the acronym \emph{``ECAI''} can be invented to turn this abstract point into a specific semantic point.

\section{Assessing the Memory Ability of RNNs in Semantic Euclidean Space}
\label{sec:rnn}

Let a set of sequences be denoted as $\mathcal{X}= \{X_1, X_2, ..., X_S\}$, where the size of  $\mathcal{X}$ is $S$, the length of each sequence in $\mathcal{X}$ is $T$, and $X_i = x_{i1}, x_{i2}, ..., x_{iT}$ represents a sequence in $\mathcal{X}$ \footnote{In this paper, the notation $(x_i)_T$ is equivalent to $x_{iT}$. Both of them denote the $T$-th element in $X_i$}. A sub-sequence of $X_i$ is represented by $(X_i)_m^n=(x_i)_m, (x_i)_{m+1}, ..., (x_i)_{n}, \text{ s.t. } m < n$. 

RNNs process and memorize a sequence of information by continuously converting the original sequence $X_i$ into the hidden layer sequence $H_i$ through recurrent units. The process is formulated as below. $F$ represents any recurrent unit, such as VRNN, LSTM and GRU.
\begin{align}
    h_{ij} = F(x_{ij}, h_{i(j-1)})
\end{align}
Specifically, an input sequence $X_i $ is composed of $n$-dimensional word embedding vectors, while the points in $X_i$ are mapped onto $\mathbb{SR}^n$ as specific semantic points. The hidden sequence $H_i$ is produced by recurrent units usually through some nonlinear functions such as $tanh$ or $sigmoid$. Therefore, it is difficult to guarantee that $H_i$ and $X_i$ are in the same space. To tackle this problem, the tied model \cite{tied-1,tied-2,tied-in-one} is applied here. It performs a dot product between each point in $H_i$ and all specific semantic points in $\mathbb{SR}^n$ where $X_i$ belongs. This ensures $H_i$ and $X_i$ are in the same space. The vectors in $H_i$ are usually abstract semantic points in $\mathbb{SR}^n$. 

It is common that the hidden layer sequence, instead of the original input sequence, is used to perform natural language processing tasks. The hidden layer information learned by RNNs carries the semantic information of the original input sequence and  meets the needs of specific tasks. The sequence order information is implicitly encoded by the recurrent structure, where a hidden unit is not a word representation by itself, but the word’s contextual representation including the neighboring words and sequence information. Therefore, the comparison between the set of input word embeddings and the set of hidden units is a spontaneous choice that reflects the capacity of memorizing the sequence information. We can assess the memory ability of RNNs by measuring the semantic relationship between the original input sequence and the hidden layer sequence. In order for a more comprehensive assessment, we propose six evaluation indicators of memory ability based on different ways of using the hidden layers of RNNs.

\subsection{Different Ways of RNNs Memorizing Sequence Information}
\label{subsec:different_way}

The hidden layers of RNNs can be used in various ways, which can be classified into two categories.

\textbf{Single Hidden Layer Representation (SHLR)} Only the last hidden layer vector $h_{it}$ is used to represent the central idea of the original input sequence $X_i$. The typical Encoder-Decoder model \cite{encoder-decoder-2} and many language models \cite{singleway-1,character-level-language-model} are examples of using this method.

\textbf{Multiple Hidden Layer Representation (MHLR)} All hidden layer vectors $H_i$ are used to represent the meaning of the original input sequence $X_i$. In other words, the meaning of $X_i$ is represented by the meaning of $H_i$. The multiway attention network (MwAN) \cite{multiway-1} and hierarchical attention networks (HAN) \cite{multiway-2} are examples of using this method.

Both representations are based on the assumption that the meaning of the original input sequence is preserved in the hidden layers learned by RNNs. Since $\mathbb{SR}^n$ provides a basis for measuring the meaning and central idea of a sequence of semantic points, in order to assess the memory ability of RNNs, we can measure the semantic difference between the original sequence and the hidden layer sequence as the spatial difference between their convex hull representations of meaning and central idea.

The memory ability of RNNs is also influenced by the input sequence length. Suppose that an evaluation indicator $\Theta$ is used to measure the memory ability of RNNs. $\Theta(W)$ denotes the memory ability of RNNs for a sequence which is of length $W$, under the evaluation indicator $\Theta$. $\langle \Theta(1), \Theta(2), ..., \Theta(T) \rangle$ reflect the change of memory ability of RNNs with the increase of sequence length under $\Theta$. 

Given an input sequence $X_i$ of length $T$, let $\mathcal{Y}$ denote the set of sub-sequences of length $W$ in each $X_i$.
\begin{equation}
\begin{split}
	\mathcal{Y}=\{(X_i)_p^{p+W-1} \big| &\forall X_i \in \mathcal{X}, \\
	&p=1, 2, ..., T-W+1\}
\end{split}
\end{equation}

$\Theta(W)$ is then the average of the memory ability over all sub-sequences in $\mathcal{Y}$ under the indicator $\Theta$.

Based on SHLR and MHLR, six evaluation indicators are defined to measure the memory ability of RNNs with different types of recurrent units from a spatial perspective. They are detailed below.

\subsection{Semantic Coverage Recall Ratio}

From the MHLR point of view, the hidden layers of RNNs should always construct its expression around the meaning of its input sequence $(X_i)_p^q$. The meaning of the hidden layer sequence $(H_i)_p^q$ should intersect with the meaning of $(X_i)_p^q$ as much as possible. Such intersection captures the valid part of the sequence information learned by RNNs, which is called Semantic Coverage and denoted as $\text{SC}$:
\begin{align}
	\text{SC}((X_i)_p^q, (H_i)_p^q) &= \text{ME}((X_i)_p^q) \cap \text{ME}((H_i)_p^q)
\end{align}

In Figure \ref{fig:duo_convex}, the shaded pink portion represents a semantic coverage. The percentage of SC in the meaning of $(X_i)_p^q$ is called Semantic Coverage Recall Ratio ($\text{SCRR}$). In order to observe the variation of SCRR with the increase of sequence length, the calculation process of $\text{SCRR}(W)$ is as follows:
\begin{align}
    \text{SCRR}(W) = \frac{\sum\limits_{i=1}^{S}{\sum\limits_{p=1}^{T-W+1} \frac{{\text{SC}((X_i)_p^{p+W}, (H_i)_p^{p+W})}}{\text{ME}((X_i)_p^{p+W})}}}{S(T-W+1)}
\end{align}

The normalization factor, $S(T-W+1)$, is the total number of subsequences in the processed sequence, so that the final metric is the average of the indicators for all subsequences processed by RNNs.

\subsection{Semantic Coverage Precision Ratio}

Like SCRR, the percentage of SC in the meaning of hidden layer sequence $(H_i)_p^q$ is called Semantic Coverage Precision Ratio ($\text{SCPR}$):
\begin{align}
			\text{SCPR}(W) = \frac{\sum\limits_{i=1}^{S}{\sum\limits_{p=1}^{T-W+1} \frac{{\text{SC}((X_i)_p^{p+W}, (H_i)_p^{p+W})}}{\text{ME}((H_i)_p^{p+W})}}}{S(T-W+1)}
\end{align}


\subsection{Semantic Coverage F-Measure}

SCPR and SCRR sometimes give contradictory indications. As a trade-off, the F1 measure takes into account both SCPR and SCRR can be used:
\begin{align}
			\text{SCFM}(W) = \frac{2*\text{SCRR}(W)*\text{SCPR}(W)}{\text{SCRR}(W)+\text{SCPR}(W)}
\end{align}

\subsection{Explicit Central Idea Offset Ratio}

After $(X_i)_p^q$ is processed by RNNs, from the SHLR point of view,  ${h_i}_q$ can be seen as an approximation of the central idea of  $(X_i)_p^q$. Therefore, the offset between the central idea of the original sequence $(X_i)_p^q$ and ${h_i}_q$ can reflect the memory ability of RNNs. Since ${h_i}_q$ is directly used as the central idea. This evaluation indicator is called Explicit Central Idea Offset, denoted by $\text{ECIO}$ and formulated as follows: 
\begin{align}
			\text{ECIO}((X_i)_p^q, h_{iq}) &= \left \| \text{CI}((X_i)_p^q), h_{iq} \right \| 
\end{align}

In Figure \ref{fig:duo_convex}, the green dotted line represents an explicit central idea offset. When an RNN processes the sequence set $\mathcal{X}$, we can use the maximum value of the sub-sequence offsets as the normalization factor, and then calculate the ratio of the average offset distance in the sub-sequence set to the maximum distance to obtain an Explicit Central Idea Offset Ratio (ECIOR).

In order to observe the variation of ECIOR with the increase of sequence length, the calculation process of $\text{ECIOR}(W)$ is as follows:
\begin{align}
	\frac{\frac{1}{S(T-W+1)} \sum\limits_{i=1}^{S}{\sum\limits_{p=1}^{T-W+1} {\text{ECIO}((X_i)_p^{p+W}, (h_i)_{p+W})}}}{\max \limits_{\substack{p=1,...,T-W+1\\
				i=1, 2, ..., S
			}}(\text{ECIO}((X_i)_p^{p+W}, (h_i)_{p+W}))}
\end{align}

\subsection{Implicit Central Idea Offset Ratio}

The meaning of the original sequence $(X_i)_p^q$ can be expressed by the meaning of the corresponding hidden sequence $(H_i)_p^q$ when using MHLR. In this case, the central idea of $(H_i)_p^q$ needs to be calculated first when considering the offset distance between the central idea of $(H_i)_p^q$ and the central idea of $(X_i)_p^q$. This evaluation indicator is called Implicit Central Idea Offset, denoted as ICIO and formulated as below:
\begin{align}
			\text{ICIO}((X_i)_p^q), (H_i)_p^q)) &= \left \| \text{CI}((X_i)_p^q), \text{CI}((H_i)_p^q) \right \|
\end{align}

In Figure \ref{fig:duo_convex}, the solid yellow line represents an implicit central idea offset. Similar to ECIOR, in order to observe the variation of Implicit Central Idea Offset Ratio (ICIOR) with the increase of sequence length, the calculation process of $\text{ICIOR}(W)$ is as follows:
\begin{align}
	\frac{\frac{1}{S(T-W+1)}\sum\limits_{i=1}^{S}{\sum\limits_{p=1}^{T-W+1} {\text{ICIO}((X_i)_p^{p+W}, (H_i)_p^{p+W})}}}{\max \limits_{\substack{p=1,...,T-W+1\\
				i=1, 2, ..., S
			}}(\text{ICIO}((X_i)_p^{p+W}, (H_i)_p^{p+W}))}
\end{align}

 The normalization factor in ECIOR and ICIOR is the maximum possible value of ECIR and ICIO in the process of processing sequence by RNNs. It makes the ECIOR and ICIOR values fall between 0 and 1, facilitating horizontal comparisons between different indicators.

\subsection{Central Idea Hit Ratio}
From the SHLR perspective, $h_{iq}$ is used to represent the central idea of $(X_i)_p^q$, thus it should be included in the meaning of $(X_i)_p^q$. Here, the inclusion of $h_{iq}$ in the meaning of $(X_i)_p^q$ is called the ``hit of central idea'', and the exclusion of $h_{iq}$ in the meaning of $(X_i)_p^q$ is called the ``miss of central idea''. Accordingly,  we define Central Idea Hit ($\text{CIH}$) as follows:	
\begin{align}
		\text{CIH}((X_i)_p^q, h_{iq}) = \begin{cases}
				1, &h_{iq} \in \text{ME}((X_i)_p^q) \\
				0, &h_{iq} \notin \text{ME}((X_i)_p^q)
			\end{cases}
\end{align}
Then the last evaluation indicator, namely Central Idea Hit Ratio (CIHR), is given as below:
\begin{align}
		\text{CIHR}(W) = \frac{\sum\limits_{i=1}^{S}{\sum\limits_{p=1}^{T-W+1} {\text{CIH}((X_i)_p^{p+W}, (h_i)_{p+W})}}}{S(T-W+1)}
\end{align}

\begin{figure*}[!htbp]
    \subfigure[PTB]
    {  
        \centering 
        \label{fig:subfig:a}
        \includegraphics[width=\textwidth]{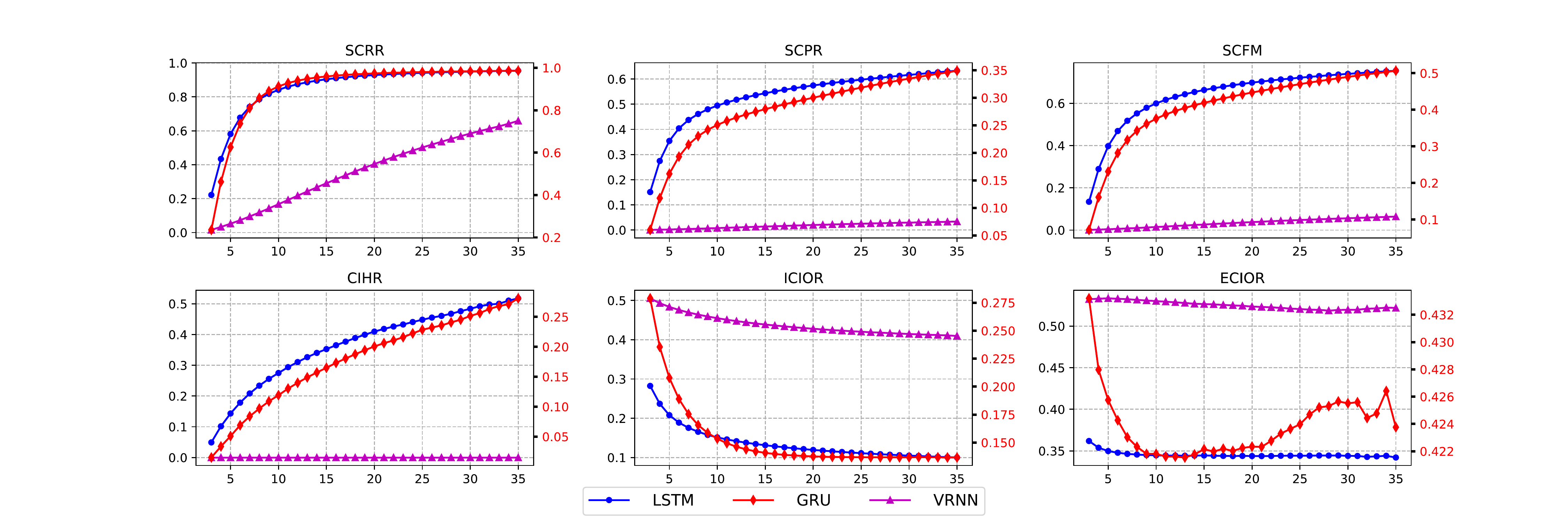}
    }
    \subfigure[Wiki-2]
    {  
        \centering 
        \label{fig:subfig:b}
        \includegraphics[width=\textwidth]{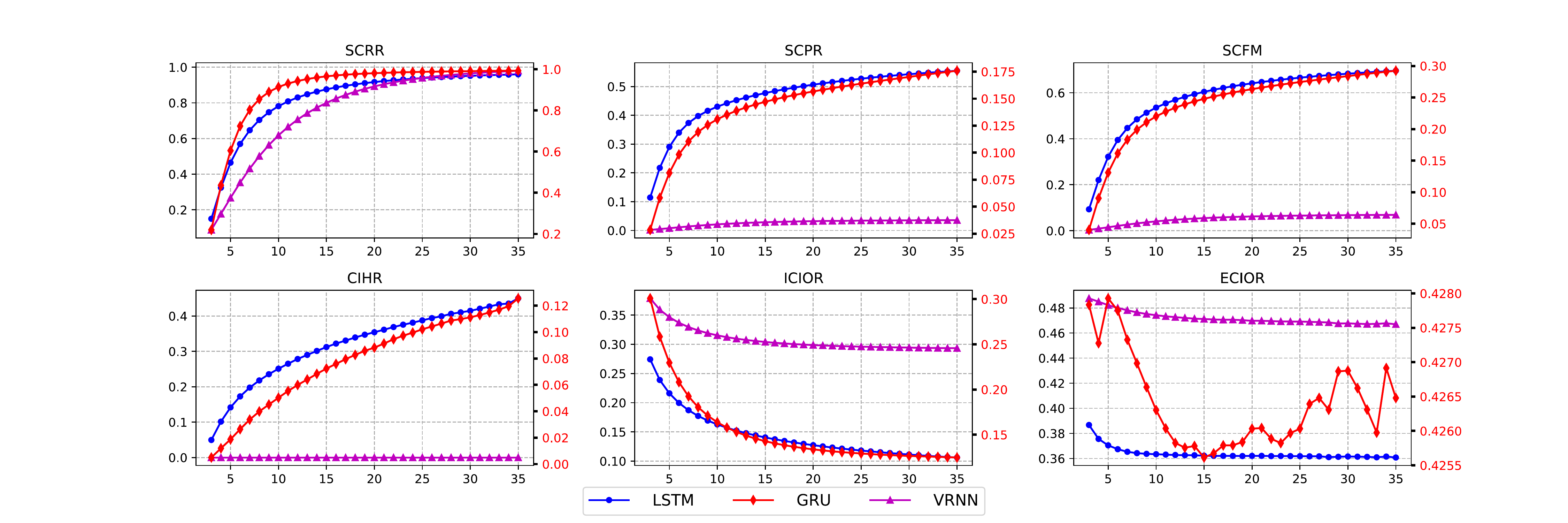}
    }
	\caption{The variation in memory ability, with respect to sequence length. The horizontal coordinates represent different settings of sequence length, and the longitudinal coordinates represent different evaluation indicators. The ordinate scale of the LSTM and RNN is shown on the left side of each picture. The ordinate scale of the GRU is on the right side of each picture.}
	\label{fig:35}
\end{figure*}

\begin{figure*}[!htp]
    \subfigure[Wiki-2]
    {  
        \centering 
        \label{leida:subfig:a}
        \includegraphics[width=0.9\textwidth]{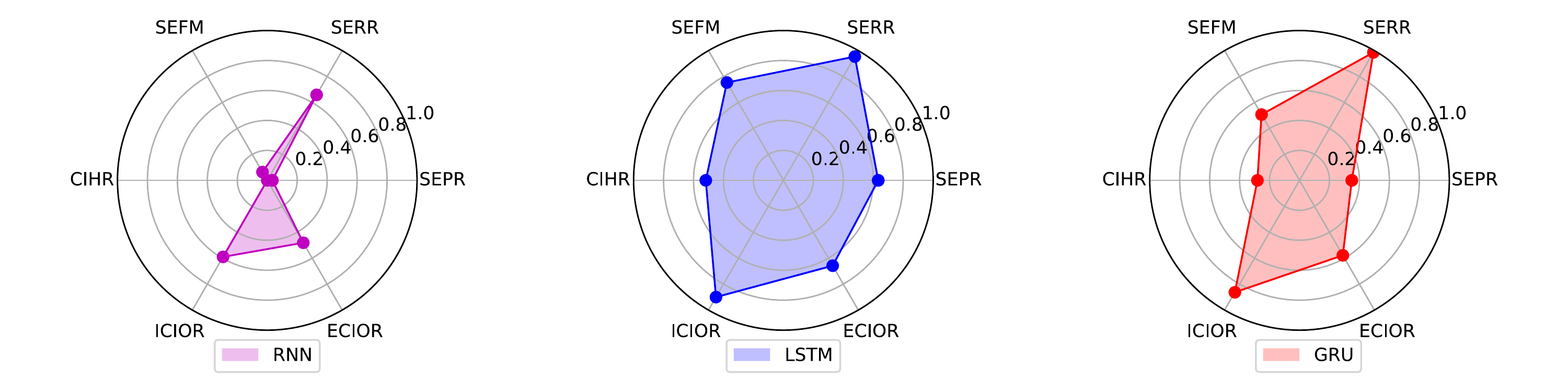}
    }
    \centering
	\label{fig:leida}
\end{figure*}
\addtocounter{figure}{-1}
\addtocounter{figure}{1}
\begin{figure*}[!htp]
    \subfigure[Wiki-2]
    {  
        \centering 
        \label{leida:subfig:b}
        \includegraphics[width=0.9\textwidth]{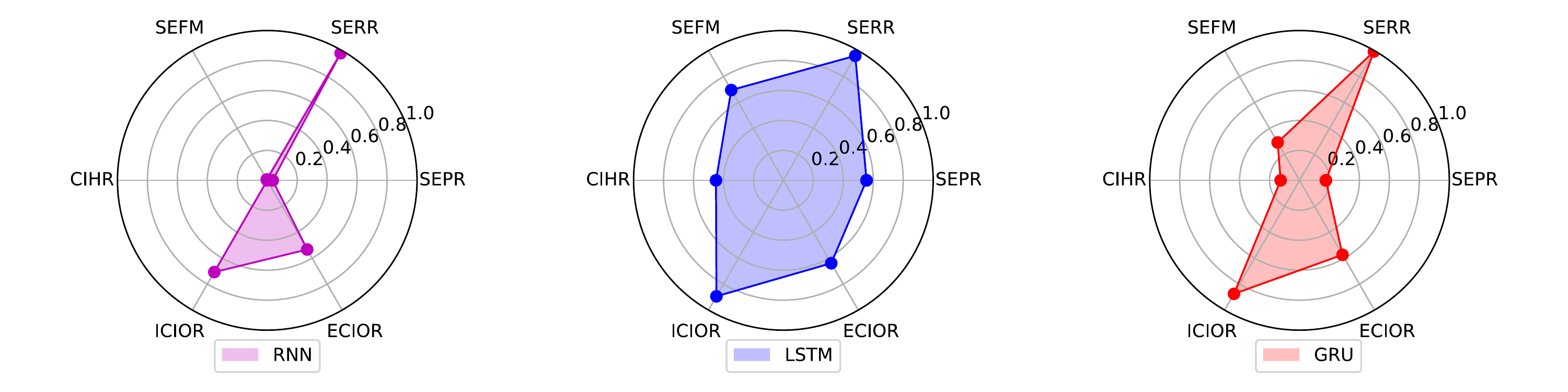}
    }
    \centering
	\caption{Comparison of memory abilities between different types of recurrent units. Each indicator takes the optimal value across different sequence lengths. Different from SCRR, SCPR, SCFM and CIHR for which the higher value indicates a better memory ability,  for ICIOR and ECIOR, the lower value the better. Thus the value corresponding to ICIOR and ECIOR shown in each picture is 1 minus the minimum of original value.}
	\label{fig:leida}
\end{figure*}

\section{Experiments}

Using the proposed six indicators, we assess the memory ability of three types of recurrent units, including VRNN ($tanh$ is used as the activation function), LSTM and GRU. Each of them is incorporated in a tied NNLM \cite{tied-1} with a single-layer RNN to perform the language modeling task on Penn Treebank \cite{ptb} (PTB) and Wiki-2 \cite{wiki2} datasets. Perplexity (PPL) is used as the performance metric. 
For the tied NNLM, the number of neurons in the hidden layer is set to 50 and BPTT is set to 35. The latter means that the long text sequences in the dataset are split into short ones with a maximum length 35, and the short sequences will further generate sub-sequences at lengths no greater than 35. These sub-sequences will constitute the set $\mathcal{Y}$ defined in Section \ref{subsec:different_way}. Moreover, the tied NNLM uses a $50$-dimensional word embedding, based on which a Semantic Euclidean Space $\mathbb{SR}^{50}$ is constructed. The hidden layer vectors also belong to $\mathbb{SR}^{50}$ in order to use the evaluation indicators defined in Section \ref{sec:rnn} to analyze the memory ability of RNNs.

The training of tied NNLM uses the stochastic gradient descent algorithm. The learning rate is initially set to be $\alpha=20$, and is halved if no significant improvement is observed on the log-likelihood of validation data. 

It should be pointed out that for calculating the indicators related to semantic coverage (SCRR, SCPR, SCFM), we can naturally consider using the area to complete the calculation of these indicators. However, it is difficult to calculate the area in high-dimensional space at present. To do so, we reduce the high-dimensional vector to $2$-dimensions through $t$-SNE \cite{t-sne}, which largely preserves the topological relationship of points in high-dimensional spaces.

\subsection{Experimental Results and Analysis}

The results for the six indicators and the variation of results with respect to the increase of sequence length, on the validation sets of PTB and Wiki-2, are shown in Figure \ref{fig:35}.

\begin{itemize}[leftmargin=*]

\item \textbf{SCRR} Both LSTM and GRU well preserve the meaning of the original sequence. The SCRR of LSTM and GRU do not increase much after the sequence length reaches 15. In contrast, a trend of non-converging growth is observed for VRNN on PTB.

\item \textbf{SCPR} When the sequence length is less than 15, the LSTM and GRU have a rapid increase in the value of SCPR, along the increase of sequence length, but then the growth rate significantly drops. However, no matter how long the sequence is processed by VRNN, its SCPR value is always less than 1\%. Combined with the SCRR of VRNNs, it can be seen that although the hidden layer of VRNN can express the semantics of the original sequence to some extent, this coverage of the original sequence semantics is obtained at the cost of reducing SCPR, which is not desirable.

\item \textbf{SCFM} SCFM has almost the same trend as SCRP. This shows that in term of semantic coverage, LSTM is slightly better than GRU, and VRNN has a relatively lower capability of semantic coverage.

\item \textbf{CIHR} Although the central idea learned by LSTM and GRU under SHLR can hit the meaning of the original sequence, a rather low hit rate is observed. The central idea learned by VRNN almost fails to hit the meaning of the original sequence.

\item \textbf{ICIOR} The central idea learned by GRU and LSTM under MHLR can well restore the central idea of the original sequence, and the ICIOR of LSTM is almost 0.1. The central idea that VRNN has learned is far from the central idea of the original sequence.

\item \textbf{ECIOR} The central idea of sequence that LSTM learns under SHLR effectively restores the central idea of the original sequence. GRU is less effective than LSTM when learning short sequences, but when the sequence length is around 15, GRU has achieved similar or even better results compared with LSTM. VRNN always maintains a large deviation from the central idea of the original sequence.

\end{itemize}

\begin{figure}[!tbp]
	\centering
	\includegraphics[width=0.7\linewidth]{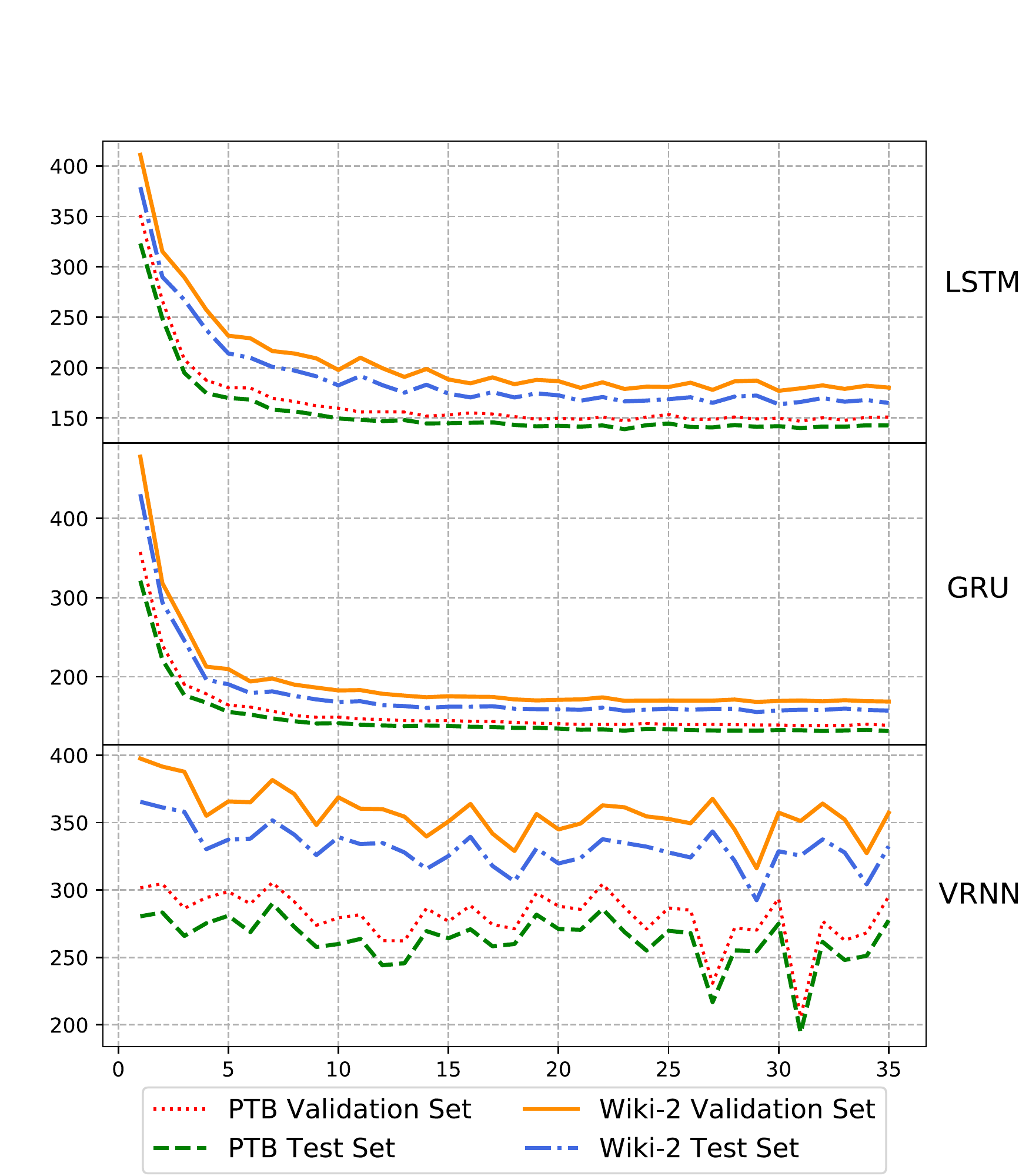}
	\caption{The influence of BPTT on PPL. The horizontal coordinates represent the BPTT, and the longitudinal coordinates represent PPL.}
	\label{fig:epoches}
\end{figure}

For SCRR, SCPR, SCFM and CIHR, a higher value indicates a better memory ability. As is shown in Figure \ref{fig:35}, the values of SCRR, SCPR and SCFM increase rapidly with the increase of the sequence length at first, and then stabilize after reaching a certain level. This to some extent reflects the process of semantic cognition. With the continuous input of words, the meaning of a sentence gradually becomes clear. The hidden layer sequence of RNNs gradually remembers the meaning of the original sequence. When the sentence reaches a certain length, the memory ability of recurrent units reaches its maximum capacity, so the information carried by the new-coming words cannot be effectively memorized by the hidden layer. 

CIHR does not show a converging trend with the increase of sequence length. We speculate that it is because the increase in the length of the sentence expands the area surrounded by the original sequence in $\mathbb{SR}^n$, which increases the probability of hits.

ICIOR and ECIOR are distance-based indicators, for which the lower value indicates a better memory ability. The trends of these two indicators are also consistent with the process of semantic cognition.


\subsection{Summary and Verification}

The proposed evaluation indicators for memory ability of RNNs turn out to be robust on both datasets. They can reliably reflect the difference in memory abilities caused by the internal factor from different perspectives.

In order for a more intuitive understanding, we plot a radar map of memory abilities of VRNN, LSTM and GRU on PTB and Wiki-2 in Figure~\ref{fig:leida}. It turns out that the memory abilities of GRU and LSTM are stronger than VRNN in all aspects. LSTM achieves better results when using Single Hidden Layer Representation (SHLR) to represent the central idea of the original sequence. When using Multiple Hidden Layer Representation (MHLR) to represent the meaning of the original sequence, LSTM shows a memory ability similar to GRU's. Moreover, all three recurrent units exhibit low values in CIHR, restraining the memory ability in this regard. This suggests that more attention could be paid to CIHR when designing new recurrent units for improved memory ability.

Finally, it is worth noting that the indicator values for both LSTM and GRU remain stable after the sequence length reaches 15. We conjecture that setting the value of BPTT to 35 (as external factor) may be unnecessary. When the value of BPTT is around 15, the PPL should be able to achieve a good performance.

In order to verify the above conjecture, we train the language model under different BPTT settings, and compute the PPL on the validation set and the test set. In Figure \ref{fig:epoches}, we find that the LSTM and GRU have achieved good results when the value of BPTT is around 15, while the PPL value of VRNN keeps fluctuating. At this point, our conjecture has been verified.

\section{Conclusions and Future Work}

In this paper, we have analyzed the internal and external factors affecting the memory ability of RNNs, and constructed a Semantic Euclidean Space ($\mathbb{SR}^n$) in which the meaning and central idea expressed by a text sequence can be measured. Further, we define six evaluation indicators on $\mathbb{SR}^n$ according to different ways of using the hidden layer of RNNs. These indicators are used for analyzing the memory ability of RNNs. Through the analysis, we are able to reveal the advantages and disadvantages of three widely used types of RNNs: VRNN, LSTM and GRU. The results give insights on the choice of recurrent unit in different scenarios, and also provide strategies for selecting the hyper-parameters of sequence length.

Future research will be focused on defining more evaluation indicators in $\mathbb{SR}^n$ to investigate the internal mechanism of RNNs, as well as applying our indicators to more complicated tasks, such as reading comprehension and machine translation. At the same time, we will explore the design of new recurrent units based on the evaluation indicators, to improve the memory ability of RNNs.

\ack This work is funded in part by the National Key Research and Development Program of China (grant No. 2018YFC0831704) and Natural Science Foundation of China (grant No. U1636203, U1736103).

\bibliography{ecai}
\end{document}